# Splenomegaly Segmentation using Global Convolutional Kernels and Conditional Generative Adversarial Networks

Yuankai Huo*[a], Zhoubing Xu[a], Shunxing Bao[b], Camilo Bermudez[c], Andrew J. Plassard[b], Jiaqi Liu[b],
Yuang Yao[b], Albert Assad[d], Richard G. Abramson[e], Bennett A. Landman[a,b,c,e]

[a] Electrical Engineering, Vanderbilt University, Nashville, TN, USA 37235
[b] Computer Science, Vanderbilt University, Nashville, TN, USA 37235
[c] Biomedical Engineering, Vanderbilt University, Nashville, TN, USA 37235
[d] Incyte Corporation, Wilmington, DE, USA 19803
[e] Radiology and Radiological Science, Vanderbilt University, Nashville, TN, USA 37235

## ABSTRACT

Spleen volume estimation using automated image segmentation technique may be used to detect splenomegaly (abnormally enlarged spleen) on Magnetic Resonance Imaging (MRI) scans. In recent years, Deep Convolutional Neural Networks (DCNN) segmentation methods have demonstrated advantages for abdominal organ segmentation. However, variations in both size and shape of the spleen on MRI images may result in large false positive and false negative labeling when deploying DCNN based methods. In this paper, we propose the Splenomegaly Segmentation Network (SSNet) to address spatial variations when segmenting extraordinarily large spleens. SSNet was designed based on the framework of image-to-image conditional generative adversarial networks (cGAN). Specifically, the Global Convolutional Network (GCN) was used as the generator to reduce false negatives, while the Markovian discriminator (PatchGAN) was used to alleviate false positives. A cohort of clinically acquired 3D MRI scans (both T1 weighted and T2 weighted) from patients with splenomegaly were used to train and test the networks. The experimental results demonstrated that a mean Dice coefficient of 0.9260 and a median Dice coefficient of 0.9262 using SSNet on independently tested MRI volumes of patients with splenomegaly.

## 1. INTRODUCTION

Spleen volume estimation is essential in detecting splenomegaly (abnormal enlargement of the spleen), which is a clinical biomarker for spleen and liver diseases [1, 2]. Manual tracing on medical images has been regarded as gold standard of spleen volume estimation. To replace the tedious and time consuming manual delineation, many previous works have been proposed to perform automatic spleen segmentation on ultrasound [3-5], computed tomography (CT) [6-10] or magnetic resonance imaging (MRI) [11-14]. In recent years, deep learning methods have shown their advantages on automatic spleen segmentation compared with traditional medical image processing methods [15]. However, the existing deep learning methods are typically deployed on CT images collected from healthy populations (e.g., spleen size < 500 cubic centimeter (cc)). When dealing with splenomegaly MRI segmentation (e.g., spleen size > 500 cc), we need to overcome two major challenges: (1) the large inhomogeneity on intensities of clinical acquired MR images (e.g., T1 weighted (T1w), T2 weighted (T2w) etc.), and (2) the large variations on shape and size of spleen for splenomegaly patients [13]. Recently, global convolutional network (GCN) have shown advantages in sematic segmentation on natural images with large variations by using larger convolutional kernels [16]. Meanwhile, adversarial networks have proven able to refine the semantic segmentation results [17].

In this paper, we propose a new Splenomegaly Segmentation Network (SSNet) to perform the splenomegaly MRI segmentation under the image-to-image framework with the end-to-end training. In SSNet, the GCN is used as the generator while the conditional adversarial network (cGAN) is employed as the discriminator [18]. To evaluate the performance of SSNet, the widely validated Unet [19] and GCN were employed as benchmark methods. Sixty clinical acquired MRI scans (32 T1w and 28 T2w) were used as the experimental cohort to test the robustness of the proposed SSNet on the multi-contrast scenario. The experimental results demonstrated that the SSNet achieved more accurate and more robust segmentation performance compared with benchmark methods.

## 2. METHODS

The SSNet was designed under the GAN framework, which consisted of both a generator and discriminator (Figure 1). In this section, we introduce each component in the SSNet.

**2.1 Generator of SSNet**

The GCN was employed as the generator in SSNet for the image-to-image segmentation, where the input and output images had the same resolution $512 \times 512$. Each training image was sent to a convolutional layer (kernel size = 1, channels = 64, stride = 2, padding = 3). Then, the "encoder" portion (left side of GCN) extracted the feature maps from the convolutional layer using four hierarchical residual blocks, which were the same as the ResNet [20]. Then, five GCN units [16] were used to transfer the feature maps for each layer to two channels using the large convolutional kernels. The equivalent kernel size was the resolution of the feature map by assembling two 1D orthogonal kernels [16]. The new feature maps with large reception field were further sent to the boundary refinement layer that is defined in [16]. Next, the refined feature maps were added to the up-sampled feature maps from the "decoder" portion (right side of GCN). Finally, the added maps were further refined by boundary refinement layer and deconvolved to the final segmentations. In Figure 1, the number of channels of each encoder was shown in the green boxes, while the number of channels of each decoder was two. The image resolution for each level was shown on the left side of Figure 1.

**2.2 Descriminator of SSNet**

In SSNet, the conditional GAN (cGAN) was used to further refine the segmentation results in the end-to-end training[18]. Briefly, estimated segmentation, manual segmentation and input images were used under the conditional manner. For the true segmentation, the ground truth for the cGAN was "true." For the segmentation from the generator, the ground truth for the cGAN was "false." The PatchGAN [18] was used as the classifier for the cGAN, which was a compromise solution between classifying the whole image and classifying each pixel.

**2.3 Loss Function and Optimization**

The loss function of SSNet was defined as $Loss_{SSNet}$ in the following equation.

$$Loss_{SSNet} = Loss_{Dice} + \lambda \cdot Loss_{GAN} \qquad (1)$$

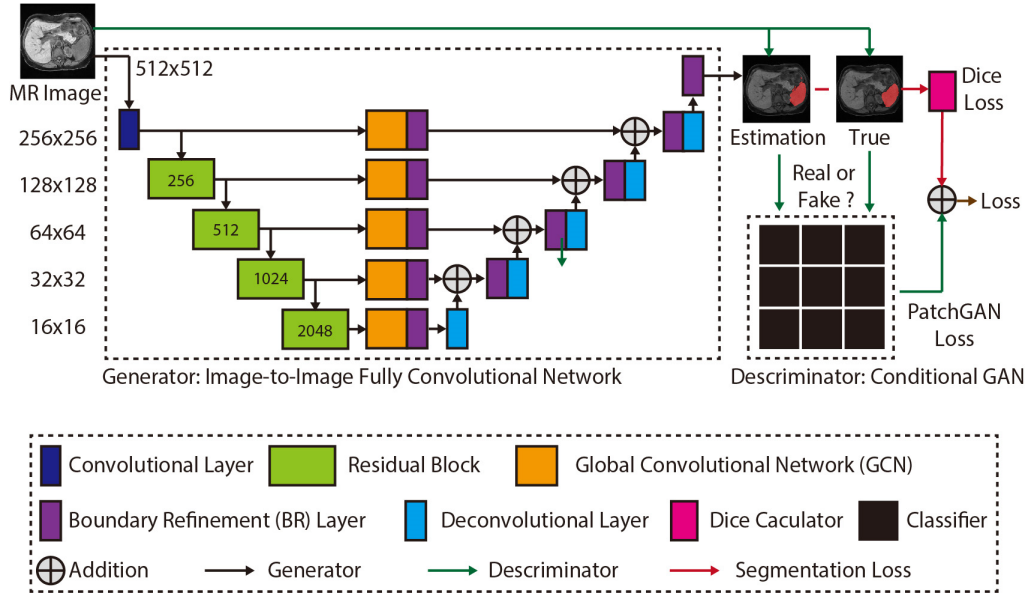

Figure 1. The proposed network structure of the Splenomegaly Segmentation Net (SSNet). The number of channels of each encoder is shown in the green boxes, while the number of channels of each decoder is two. The image (or feature map) resolution for each level is shown on the left side of this figure.

$\text{Loss}_{\text{Dice}}$ represents the Dice loss, which was the negative Dice similarity coefficient (DSC) score between the segmentation from the generator and the manual segmentation. The $\text{Loss}_{\text{GAN}}$ indicated the GAN loss, which was the binary cross entropy (BCE) loss between the cGAN estimations and true classes. The λ was a constant value that decided the weights when adding the two losses. In our study, the λ was empirically set to 100. The Adam optimization [21] was used as the optimization function (learning rate = 0.00001).

## 3. EXPERIMENTS

### 3.1 Data

We used 60 clinically acquired abdominal MRI scans (32 T1w / 28 T2w) from splenomegaly patients to evaluate the performance of different deep convolutional networks. Images were acquired after informed consent and the study was monitored by an approved institutional review board. The data accessed in this study was de-identified. Among the entire cohort, 45 scans (24 T1w / 21 T2w) were used as training data, while the remaining 15 scans (8 T1w / 7 T2w) were employed as independent validation data. For each scan, the MRI volume was resampled to 512 × 512 × 512 resolution to obtain 512 axial, 512 coronal as well as 512 sagittal 2D images. The manual segmentations of spleens were traced by an experienced rater using the Medical Image Processing Analysis and Visualization (MIPAV) software [22]. From the manual segmentations, the minimum size of spleen is 368 cubic centimeter (cc), the maximum size is 5670 cc, the mean spleen volume is 1881 cc, and the standard deviation is 1219 cc.

### 3.2 Experiments

Two sets of the experiments were performed to compare the performance of the proposed SSNet with Unet and GCN benchmarks. Since it was a 2D segmentation problem, we used the ImageNet [23] pertained model as the initialization for each network when the pertained model was available. The first set of the experiments only used the axial images as both training and testing images. Then, the 3D volumetric spleen segmentations were derived by assembling the testing images slice by slice from the same testing scan. For the second set, all axial, coronal and sagittal view 2D images from the 45 resampled training scans were used to train three networks: (1) the first network (axial view network) was trained by all axial view images, (2) the second network (coronal view network) was trained by all coronal view images, and (3) the third network (sagittal view network) was trained by all sagittal view images. In the testing procedure, the 15 independent testing scans were used for an external validation. For each resampled testing scan, all axial view 2D images were segmented by the axial view network and then concatenated to a 3D segmentation. Similarly, 3D segmentations from coronal and sagittal views for such testing scan were obtained from the coronal view network and the sagittal view network. Finally, the three 3D segmentations (for each testing scan) were fused to one final segmentation by (1) merging three segmentations from different views to a single segmentation using "union" operation, (2) performing open morphological operations to smooth the boundaries, and (3) performing close morphological operations to fill the holes.

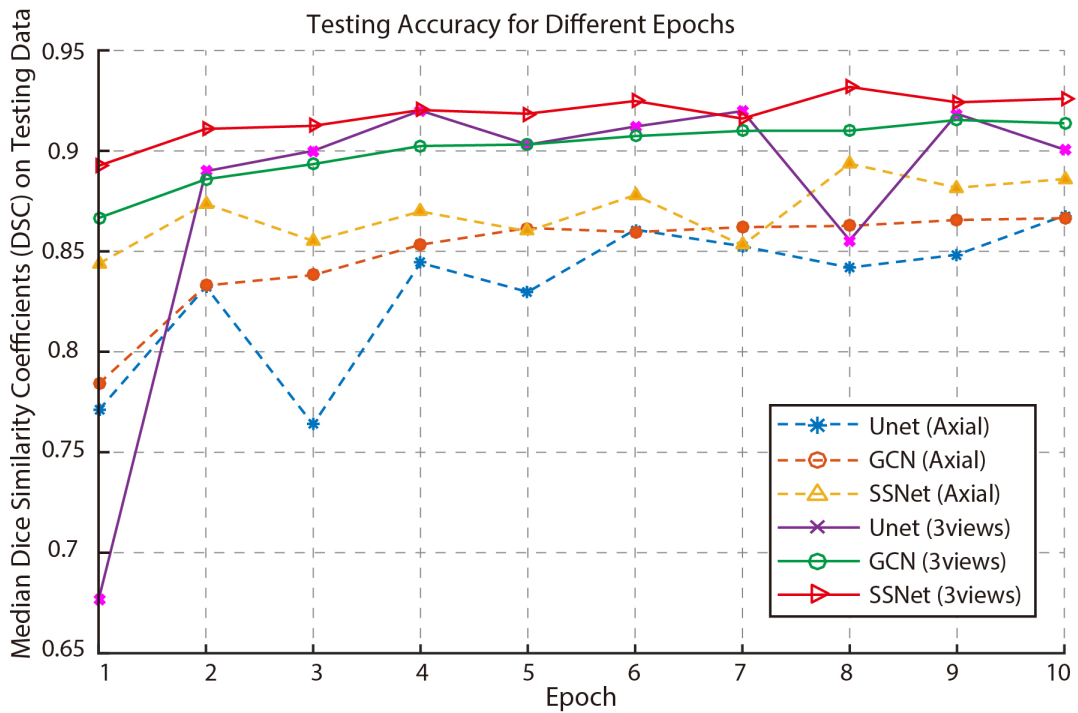

Figure 2. The testing accuracy of different epochs was shown in this figure. The y axial indicated the mean Dice similarity coefficients (DSC) on all testing volumes, while the x axial presented the epoch number from one to ten. The dashed curves were the testing accuracy for the case that only axial images were used as training and testing images. The solid curves were the testing accuracy for the case that all axial, coronal and sagittal view images were used in both training and testing scenario.

### 3.3 Validation Metrics

The Dice similarity coefficient (DSC) values relative to the manual segmentation were used as the metrics to evaluate the performance of different segmentation methods. All statistical significance tests were made using a Wilcoxon signed rank test ($p<0.01$).

## 4. RESULTS

Figure 2 presents the testing accuracy of different methods and experimental strategies as median DSC curves for ten epochs. The y axial indicated the mean Dice similarity coefficients (DSC) on all testing volumes, while the x axial presented the epoch number. The dashed curves were the testing accuracy for the case that only axial images were used as training and testing images. The solid curves were the testing accuracy for the case that all axial, coronal and sagittal view images were used in both training and testing processing. From this figure, the mean testing accuracy plots were systematically increased when trained with more epochs. For most of the epochs, the proposed SSNet achieved more accurate testing results than GCN and Unet on both single view and multi-view training scenarios.

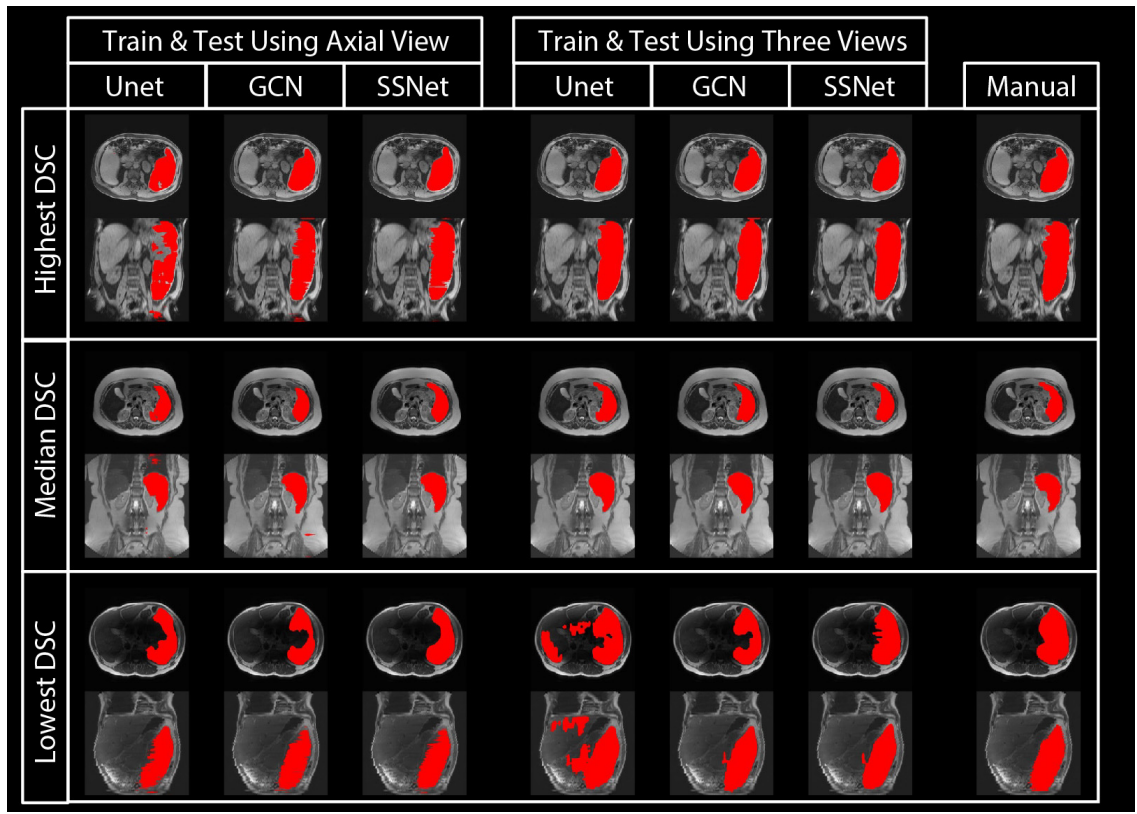

Figure 3. The qualitative results of different methods. The segmentation results of Unet, GCN and SSNet on using (1) only axial 2D images, and (2) all axial, coronal and sagittal 2D images are shown in the figure for different columns. The manual segmentation results for the same subjects are presented as well. The results of three subjects were selected from the highest, median and lowest DSC from the SSNet's testing data.

Figure 3 presents the qualitative results of different deep learning methods along with the manual segmentation. The upper, middle and lower rows were corresponding to the subjects with highest, median and lowest DSC values of SSNet using three views. The segmentation results of Unet, GCN and SSNet on using (1) only axial 2D images, and (2) all axial, coronal and sagittal 2D images were shown in the figure for different columns. The manual segmentation results for the same subjects were presented as the right-most column. In Figure 4 presents the quantitative results of different deep learning methods as box plots. All the other methods were compared with the proposed SSNet using three view images ("Ref."). The proposed method achieved significantly better DSC results ($p<0.01$) than methods with "*" except the one with "N.S.". The lowest DSC value of the SSNet is smaller than the benchmark methods. From Figure 3 and 4, the GCN outperformed the Unet by capturing the large spatial variation for the splenomegaly segmentation. By adding GAN supervision, the proposed method not only alleviated the outliers but also achieved the higher median DSC (0.9262) and mean DSC (0.9260) compared with baseline methods. Meanwhile, using richer training data on three imaging views leveraged the segmentation performance for a significant margin.

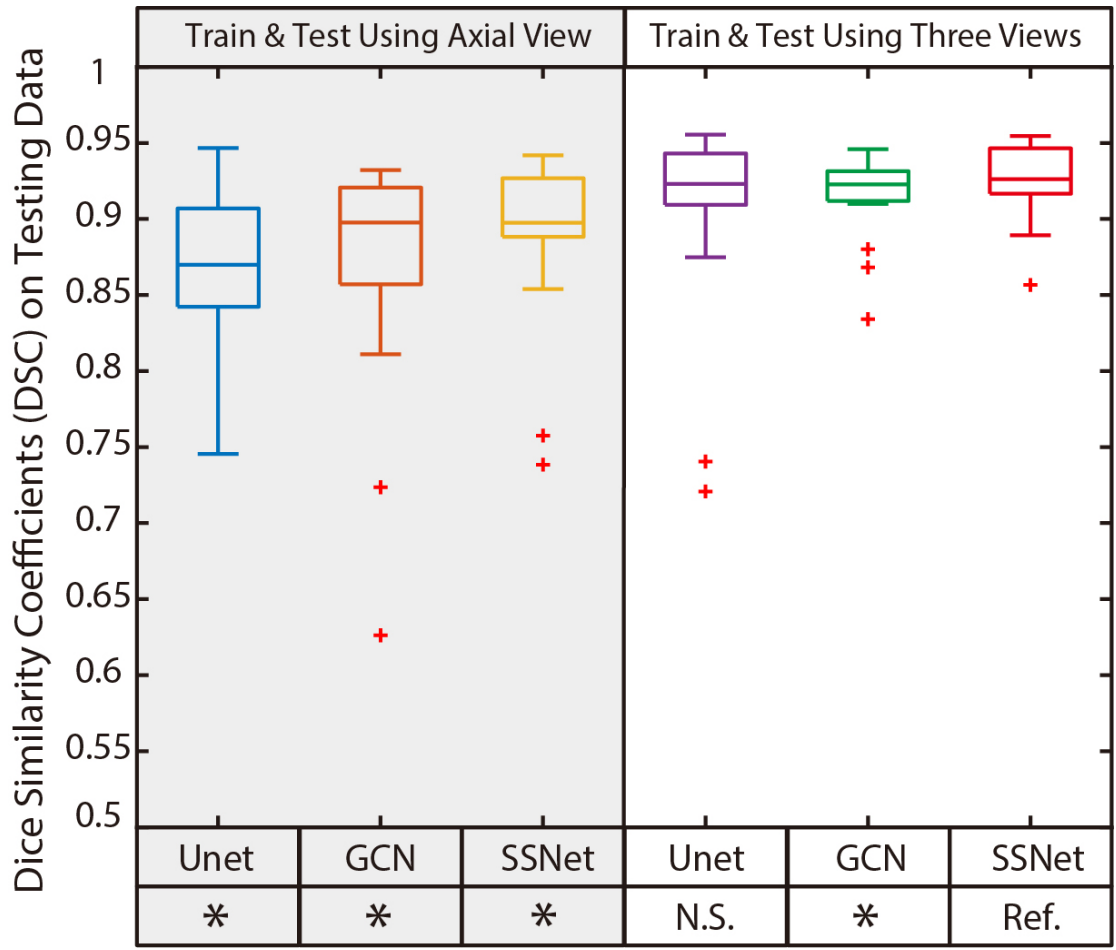

Figure 4. The quantitative results of different methods. The box plots in left panel indicate the results of using only axial view images, while the right panel presents the results of using all axial, coronal and sagittal images as in both training and testing. The Wilcoxon signed rank tests were employed as statistical analyses, where "Ref." indicates the reference method. The "*" indicates the $p<0.01$ while the "NS" means not significant.

## 5. CONCLUSION AND DISCUSSION

We proposed the SSNet to perform the splenomegaly segmentation using MRI clinical acquired scans. From the experimental results, the proposed method provided us 0.9262 and 0.9260 for the median and mean DSC, which was significantly better than the baseline methods when trained on the axial view. Richer training data in the form of 2-D triplanar sections improved all methods, but SSNet remained superior than GCN and had fewer outliers than Unet. From Figure 2, the proposed SSNet achieved generally better performance on median DSC compared with benchmark methods on different epoch numbers. From Figure 3 and 4, the SSNet was shown to achieve more accurate (higher median DSC) and more robust (higher lowest DSC) segmentation performance compared with benchmark results. The results also demonstrated that using all axial, coronal and sagittal images as both training and testing data consistently provided us better segmentation performance than using single axial view.

The major limitation of this work was that the segmentation was performed on the 2D images, which might lose the 3D spatial information. In the future, it would be worth exploring 3D deep neural networks to conduct the splenomegaly segmentation. Another interesting direction could be to integrate the clinical diagnostic information to the image segmentation using the attention models [24].

## 6. ACKNOWLEGEMENTS

This research was supported by NSF CAREER 1452485, NIH grants 5R21EY024036, R01NS095291 (Dawant), and InCyte Corporation (Abramson/Landman). This research was conducted with the support from Intramural Research Program, National Institute on Aging, NIH. This study was in part using the resources of the Advanced Computing Center for Research and Education (ACCRE) at Vanderbilt University, Nashville, TN. This project was supported in part by ViSE/VICTR VR3029 and the National Center for Research Resources, Grant UL1 RR024975-01, and is now at the National Center for Advancing Translational Sciences, Grant 2 UL1 TR000445-06. We appreciate the NIH S10 Shared Instrumentation Grant 1S10OD020154-01 (Smith), Vanderbilt IDEAS grant (Holly-Bockelmann, Walker, Meliler, Palmeri, Weller), and ACCRE's Big Data TIPs grant from Vanderbilt University. We gratefully acknowledge the support of NVIDIA Corporation with the donation of the Titan X Pascal GPU used for this research.